\documentclass[10pt,twocolumn,letterpaper]{article}

\usepackage[pagenumbers]{cvpr} 

\definecolor{cvprblue}{rgb}{0.21,0.49,0.74}
\usepackage[pagebackref,breaklinks,colorlinks,allcolors=cvprblue]{hyperref}
\usepackage{graphicx}       
\usepackage{subcaption}
\usepackage{multirow}

\def\paperID{****} 
\def\confName{{****}}
\def\confYear{2026}

\title{RT-Counter: Real-Time Text-Guided Open-Vocabulary Object Counting}

\author{{Hao-Yuan Ma, Li Zhang, Zhiwei Zhu, Jie Gao}\\
School of Computer Science and Technology, Soochow University\\
Suzhou, China
}

\begin{document}
\maketitle
\begin{abstract}
Text-guided open-vocabulary object counting (TOOC) aims to count objects belonging to the categories specified by natural language descriptions.
Although vision-language pre-trained models have been successful applied to TOOC tasks, they still struggle with fine-grained spatial understanding and real-time inference requirements in counting scenarios. 
To address these limitations, this paper proposes a real-time TOOC framework, called the Real-Time Counter (RT-Counter), that achieves not only good counting accuracy but also high computational efficiency. 
RT-Counter designs a novel Visual Prototype Textualization (VPT) module that can project learned visual features into a text feature space and then generate features containing the abstract information that is hard to capture with visual prototypes and the detailed prototype information that is difficult to describe in text, enhancing the object-level visual-language model's counting capabilities. 
Additionally, RT-Counter incorporates our Weaving Transformer (Weaformer) layers, maintaining high descriptive power at a fraction of the computational cost. The Weaformer layer adopts a novel hybrid attention mechanism that can efficiently weave together local and global visual features. 
Extensive experiments on three public datasets show that RT-Counter successfully breaks the accuracy-speed trade-off in TOOC. While achieving a competitive MAE of 13.30 on FSC147, RT-Counter operates at 112.48 FPS, making it 7.4x faster and over 4$\times$ more parameter-efficient than the existing leading methods in TOOC.
Our work aims at balancing high accuracy and real-time performance in TOOC.
Code is available at: https://github.com/Jason-Mar1/RT-Counter.
\end{abstract} 
\section{Introduction}
\label{sec:intro}

Object counting is a fundamental task of determining the number of specific objects in images, which plays a crucial role in numerous real-world applications including surveillance systems \cite{P2PNet}, autonomous driving \cite{CLIP-Count}, retail analytics \cite{FGENet}, and ecological monitoring \cite{FamNet}. 
Traditional counting paradigm typically relies on the closed-set assumption. In this paradigm, models are trained and tested on predefined object categories with abundant labeled data \cite{FGENet,MCNN,yolox}. 
However, this paradigm severely limits these models' practical applicability as real-world scenarios often involve the open-vocabulary counting task where novel object categories emerge continuously. In this case, it is impractical and expensive to collect comprehensive training data for every possible object type. 
Therefore, there has been a paradigmatic shift from closed-set to open-vocabulary object counting. The new paradigm demands models capable of understanding and counting arbitrary object categories described in natural language, without requiring extensive retraining or category-specific annotations.

\begin{figure}
    \centering
    \includegraphics[width=1\linewidth]{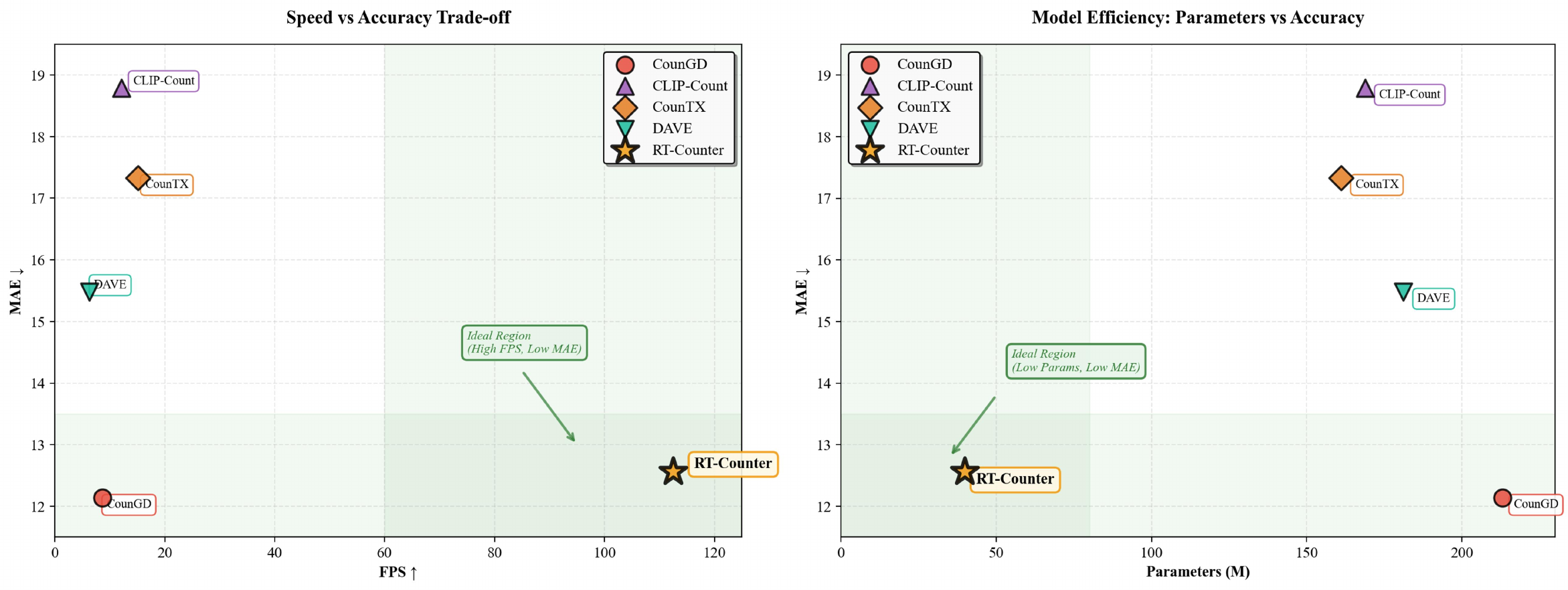}
    \caption{Comparison of different models about model counting performance vs. inference speed and model parameters, where MAE is used as the key performance metric, while FPS reflects inference speed and parameter count indicates model size.}
    \label{fig:0_realtime}
\end{figure}

The core challenges of open-vocabulary object counting lie in bridging the semantic gap between visual object prototypes and textual category descriptions as well as consistently maintaining counting accuracy across diverse object types.
Current approaches for open-vocabulary object counting can mainly be categorized into two types: visual prototype-guided and text-guided methods.
The visual prototype-guided method can be further categorized into two sub-types: few-shot and zero-shot methods.
The few-shot method learns a generalized visual prototype from some given samples for counting any object, with representative examples including FamNet \cite{FamNet}, CounTR \cite{Countr}, and LOCA \cite{loca1}. 
By contrast, the zero-shot method, such as LOCA \cite{loca1} and DAVE \cite{DAVE}, extracts the most frequent visual features as the visual prototype when counting any object. 
Thus, it is highly impossible for the zero-shot method to count rare categories.
The text-guided method predominantly relies on pre-trained image-text alignment models that serve as the core backbone for bridging visual and textual semantics in open-vocabulary object counting. Representative pre-trained models are CLIP \cite{CLIP}, Open-CLIP \cite{openclip}, and GroundingDINO \cite{GroundingDINO}. 
By leveraging the alignment capability of CLIP, researchers have proposed CLIP-Count \cite{CLIP-Count} and VL-Counter \cite{VLCounter} for object counting. Additionally, CounTX is built on Open-CLIP \cite{CounTX}, and CountGD \cite{CountGD} is developed based on GroundingDINO \cite{CountGD}. 
A critical limitation of such methods lies in their universal heavy reliance on pre-trained base models with high computational costs, which not only leads to excessively their high inference overhead but also renders them infeasible for real-time deployment scenarios.
Furthermore, the detection outputs of most GroundingDINO-based methods are non-differentiable, which makes these methods incompatible with end-to-end optimization in downstream tasks (e.g., text-to-image generation) because gradient flow is a core prerequisite for end-to-end optimization.

To address these limitations, we propose Real-Time Counter (RT-Counter), a novel framework that achieves efficient real-time text-guided open-vocabulary object counting (TOOC). 
To enable real-time performance, we design the Weaving Transformer (Weaformer) layer, a highly efficient Transformer layer that effectively weaves local and global visual features. 
To enhance feature representation, we developed a novel Visual Prototype Textualization (VPT) module. The main goal of this module is to project the learned visual features into a text feature space and generate fused features that describe not only abstract information but also detailed prototype information. 	
Extensive experimental results demonstrate that RT-Counter achieves a new benchmark while maintaining a real-time inference speed.
Our main contributions can be summarized as follows:
\begin{itemize}
    \item We propose the VPT module, a novel projection fusion module. It enriches the learned visual prototypes with abstract semantics by projecting them into the text feature space. This process creates comprehensive cross-modal representations that capture both fine-grained visual details and high-level context, leading to more accurate counting.
    \item We design the efficient Weaformer layer, which effectively weaves local and global visual features to handle objects at multiple scales. This design maintains high descriptive power while drastically reducing computational cost, enabling real-time performance.    
    \item We present RT-Counter, a real-time TOOC framework, based on VPT and Weaformer. RT-Counter employs a lightweight network architecture to extract multi-scale features from images and utilizes a feature enhancer module to capture objects of different scales within the encoded image representations. The feature enhancer module consisting of Weaformer layers iteratively refines visual features with guidance from the VPT-enriched prototypes. Therefore, RT-Counter enables efficient processing of diverse object sizes while maintaining computational efficiency for real-time counting applications.               
\end{itemize}

\section{Related Work}
\label{sec:rel}

\subsection{Visual prototype-guided counting methods} 
For visual prototype-guided counting tasks, we can use few-shot or zero-shot counting approach to capture visual prototypes.
Among few-shot methods, FamNet \cite{FamNet} is the first one that introduces few-shot counting to learn generalized visual prototypes from provided sample instances, enabling counting of arbitrary objects. 
CounTR~\cite{Countr} employs transformer architecture to encode visual exemplars and query images for class-agnostic counting.
Both FamNet and CounTR are class-agnostic approaches that introduce visual exemplar-based counting for arbitrary objects through reference images.
LOCA \cite{loca1} is a method for both few-shot and zero-shot scenarios, which utilizes ResNet to extract features and employs adaptive pooling or RoI methods to obtain prototype features. In zero-shot, LOCA extracts visual prototypes from the most frequent visual features through density-aware visual encoding without requiring labeled samples.
DAVE \cite{DAVE} is similar to the LOCA method, but it achieves object counting by clustering the feature.
These methods have their own shortcomings. Few-shot methods require manual provision of labeled samples, while zero-shot methods struggle with rare object categories owing to their reliance on frequent visual patterns. Moreover, these methods remain limited by poor generalization to novel object categories.
\begin{figure*}
    \centering
    \includegraphics[width=0.85\linewidth]{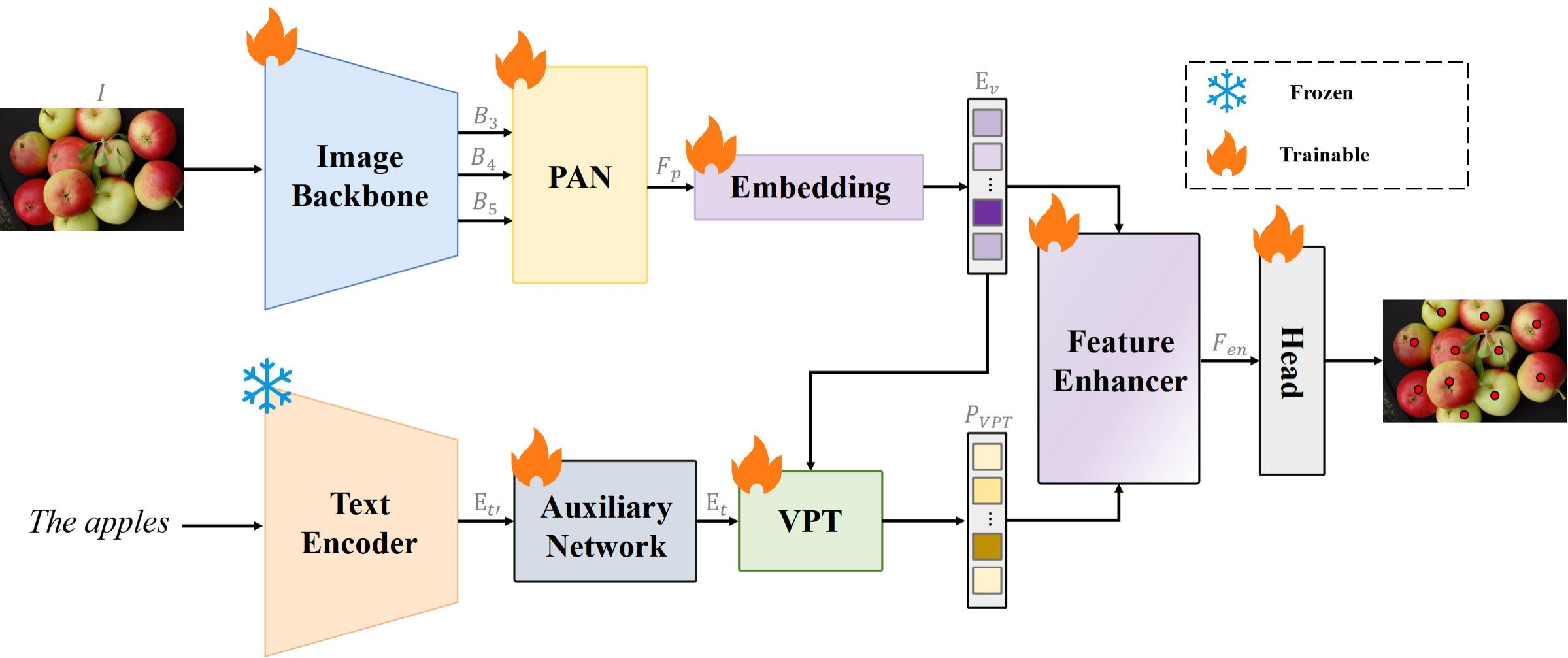}
    \caption{Overview of RT-Counter, where snowflake symbols mean that parameters are frozen, and flame symbols denote that parameters are learnable. }
    \label{fig2:arch}
\end{figure*}

\subsection{Text-guided counting methods} 
Text prompt-based methods leverage pre-trained vision-language models to bridge visual and textual semantics for open-vocabulary counting. CLIP is a common used pre-trained vision-language model. CLIP-Count \cite{CLIP-Count} first applies CLIP's image-text alignment capabilities for zero-shot counting by matching visual features with text descriptions. VL-Counter \cite{VLCounter} improves CLIP-based counting through semantic-conditioned prompt tuning and learnable affine transformation modules. CounTX~\cite{CounTX} is based on Open-CLIP with enhanced text-image fusion mechanisms for better counting accuracy. GroundingDINO is another vision-language model. CountGD employs the grounding capabilities of GroundingDINO to detect and count objects specified by descriptions in natural language, achieving promising open-vocabulary results \cite{CountGD}. Zero-shot Object Counting (ZSC) employs generative models to create representations for unseen classes guided by semantic information \cite{zsc}. VA-Count addresses zero-shot challenges through exemplar enhancement and noise suppression modules for improved visual-textual association \cite{VACount}. 
These mentioned methods suffer from heavy computational costs because of the reliance on large pre-trained models (CLIP and GroundingDINO), making them unsuitable for real-time deployment. In addition, many methods produce non-differentiable outputs that are incompatible with other tasks.

Our approach addresses the fundamental limitations of existing methods, which achieves real-time performance while maintaining open-vocabulary capabilities through our new designed VPT module. Unlike previous approaches that sacrifice computational efficiency for semantic richness or compromise open-vocabulary capabilities for speed, our method bridges this gap by efficiently compensating for the lack of detailed prototype information in text descriptions while supplementing abstract information that visual prototypes struggle to capture. Figure \ref{fig:0_realtime}  shows the scatter diagram of model counting performance vs. inference speed and model parameters.

\section{Approach}

\subsection{Pipeline Overview}
As illustrated in Figure~\ref{fig2:arch}, RT-Counter mainly comprises an image backbone, a text encoder, a Path Aggregation Network (PAN) \cite{PAN}, an embedding module, an auxiliary network, a VPT module, a feature enhancer, and a prediction head, where the parameters in the components marked with snowflake symbols are frozen, and those in components with the flame symbols are learnable. The image backbone extracts visual features from input images, while the text encoder processes the textual description. 
PAN enhances the counting performance for objects with varying sizes by fusing features from different stages of the backbone.
The embedding module is a multi-scale visual information aggregation component that primarily integrates multi-scale information, thereby enhancing the detection capability for small objects.
The auxiliary network projects initial textual representations onto the same embedding space, thus bridging the gap between visual and textual modalities.
The VPT module performs visual prototype projection and cross-modal fusion to compensate for missing detailed prototype information in text and to supplement the abstract information that visual prototypes struggle to capture. 
The feature enhancer refines the fused representations, and the final prediction head generates the counting results with precise object localization.

Given an input image $I \in \mathbb{R}^{3 \times H \times W}$ and a textual prompt $T$, where $H$ and $W$ are the height and width of input image, the goal of RT-Counter is to count the number of objects specified by the text prompt $T$ within the image $I$. Specifically, RT-Counter processes the input through the following pipeline.

The image backbone first encodes the input image $I$ and then generates multi-scale visual features $B_3$, $B_4$, and $B_5$ at different resolutions. These features are then fed into PAN to obtain aggregated features $F_p$, which are subsequently processed by the embedding module to generate visual embeddings $E_v$.
Simultaneously, the textual description $T$ is processed by the text encoder to produce initial text embeddings $E_{t'}$, which are then refined through a lightweight auxiliary network to obtain the final text embeddings $E_t$.

The core innovation of this paper lies in VPT that fuses the visual embeddings $E_v$ and the text embeddings $E_t$ to generate enhanced prototype representations $P_{VPT}$. That is, 
\begin{equation}
P_{VPT} = \text{VPT}(E_v, E_t)
\end{equation}
Further, $P_{VPT}$ is then combined with the original visual embeddings $E_v$ through the Feature Enhancer. Then, we can have enriched features $F_{en}$ as follows.
\begin{equation}
F_{en} = \text{FeatureEnhancer}(P_{VPT}, E_v)
\end{equation}

Finally, the enhanced features $F_{en}$ are passed through prediction head to generate a set of predictions $\{(\hat{x_i}, \hat{y_i}), \hat{c_i}|, i \in N_{Grid}\}$, where $N_{Grid}$ is the total number of anchor points,  $(\hat{x_i}, \hat{y_i})$ represents the predicted object location, and $\hat{c_i}$ denotes the confidence scores. Assume that the object with a confidence score exceeding a threshold $\phi$ is considered a positive detection. Thus, the number of predicted objects is given by, 
\begin{equation}
N_{Pred} = \sum_{i=1}^{N_{Grid}} \mathbb{I}(\hat{c_i} > \phi)
\end{equation}
where $\mathbb{I}(\cdot)$ is the indicator function. Let $N_{GT}$ be the number of ground truth objects. The training loss is computed based on the predicted results that are matched with ground truth annotations by using the Hungarian algorithm.

\begin{figure*}
    \centering
    \includegraphics[width=1\linewidth]{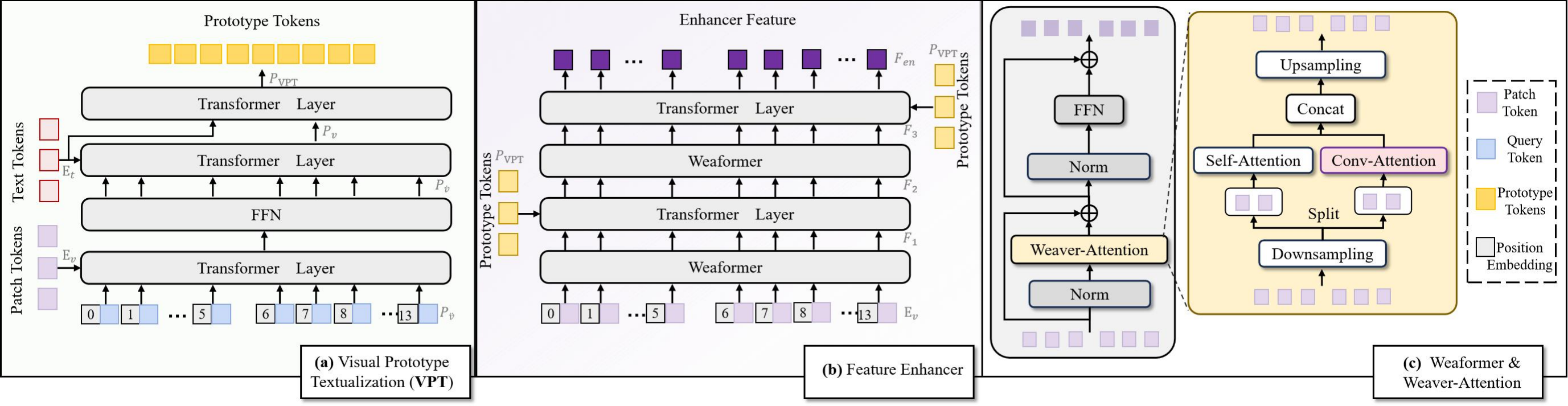}
    \caption{Detailed Architectures of VPT and feature enhancer.}
    \label{fig3:main_comp}
\end{figure*}

\subsection{Visual Prototype Textualization Module} \label{sec3_2:vpt}
Existing text-guided counting methods struggle with the inherent limitation that textual descriptions cannot capture fine-grained visual details such as texture, shape variations, and spatial arrangements, while visual prototype-guided methods fail to encode abstract semantic information like contextual relationships and categorical hierarchies. 
To address this fundamental challenge of bridging the semantic gap between visual and textual modalities, we design the VPT module that enables bidirectional information compensation—enriching text descriptions with the detailed visual prototype information while supplementing visual prototypes with the abstract semantic knowledge.

As illustrated on the left of Figure~\ref{fig3:main_comp}(a), the VPT module consists of three Transformer layers and a Feed-Forward Network (FFN). This module operates through the following stages:

\textbf{Visual Prototype Query Generation:} We first initialize a learnable visual prototype query matrix $P_{\ddot{v}} \in \mathbb{R}^{N_p \times D}$, where each row of $P_{\ddot{v}}$ is a prototype query vector, $N_p$ represents the number of prototype queries, and $D$ is the feature dimension of queries. These queries serve as anchors to extract relevant visual patterns from the input features.

\textbf{Visual Prototype Extraction and Projection:} The visual prototype query matrix $P_{\ddot{v}}$ (serving as Query) interacts with visual embeddings $E_v$ (serving as Key and Value) through a cross-attention mechanism. A refined visual prototype matrix $P_{\dot{v}}$ can be expressed as
\begin{equation}
P_{\dot{v}} = \text{FFN}(\text{Norm}(\text{CrossAttn}(P_{\ddot{v}}, E_v, E_v)))
\end{equation}
where $\text{CrossAttn}$ denotes the cross-attention operation, $\text{Norm}$ means layer normalization, and $\text{FFN}$ is the  FFN operation. It should be noted that $\text{FFN}$ project the visual prototype into the text feature space.

\textbf{Visual Prototype Filtering and Enhancement:} To filter and enhance semantic information, we further apply $\text{CrossAttn}$ to $P_{\dot{v}}$ (as Query) and $E_t$ (as Key and Value) and obtain the semantically-aware visual representations $P_v$ as follows
\begin{equation}
P_v = \text{CrossAttn}(P_{\dot{v}}, E_t, E_t)
\end{equation}
This operation filters out visual prototypes that are not semantically relevant and enriches the visual prototype with abstract textual semantics.

\textbf{Prototype Fusion:} To effectively combine the enhanced prototypes $P_v$ with the original text embeddings $E_t$, we employ a transformer layer. The inputs are first concatenated and then processed through a self-attention mechanism
\begin{equation}
P_{VPT} = \text{SelfAttn}(\text{Concat}(P_v, E_t))
\end{equation}


\subsection{Feature Enhancer and Weaformer} \label{sec3_3:fe}

In order to extract object information from visual embeddings, we design the feature enhancer module that alternates between Weaformer and Transformer layers, as shown in Figure \ref{fig3:main_comp} (b).

\textbf{Feature Enhancer:}
The feature enhancer outputs are respectively expressed as
\begin{align}
F_1 &= \text{Weaformer}(E_v) \\
F_2 &= \text{CrossAttn}(F_1, P_{VPT}, P_{VPT}) \\
F_3 &= \text{Weaformer}(F_2) \\
F_{en} &= \text{CrossAttn}(F_3, P_{VPT}, P_{VPT})
\end{align}
where $\text{CrossAttn}$ is the operation of cross-attention. 
This alternating design ensures visual features maintain spatial relationships while progressively querying the object through VPT.

\textbf{Weaformer:}
\label{sec:weaformer_def}
The standard self-attention mechanism in Transformer suffers from a quadratic computational complexity of $\mathcal{O}(N^2 C)$, where $N$ and $C$ are the numbers of input tokens and channels, making it a bottleneck for high-resolution images.
To address this, we design the Weaformer layer, a novel and efficient replacement for the standard Transformer layer, as illustrated in Figure~\ref{fig3:main_comp}(c). 

The Weaformer layer replaces the standard self-attention module with our proposed Weaver-Attention mechanism that can efficiently weave together local and global information. 
Its process begins with a downsampling operation. Let $S$ be the downsampling rate, where $S\geq 4$.
These tokens are then split in channels, strategically allocating only a small fraction (25\%) to the computationally intensive global path, while the majority (75\%) processed by the efficient local path.
The global path preserves long-range spatial relationships by applying a standard self-attention mechanism on its allocated $\frac{N}{S}$ tokens, resulting in a complexity of $\mathcal{O}\left(\left(\frac{N}{4}\right)^2 \frac{C}{4}\right)$. 
Meanwhile, the local path captures the fine-grained context using our proposed Convolutional Attention (Conv-Attention). This block generates query ($Q$), key ($K$), and value ($V$) projections using efficient DWConv and constrains the attention operation to a local window of size $k \times k$ by leveraging an \texttt{unfold} operation. This design is key to its efficiency because each token only interacts with a fixed number $k^2$ of neighboring tokens instead of $N^2$, where $k$ is a small constant, say 1 or 3. The complexity of this local path remains linear, i.e., $\mathcal{O}(\frac{N}{S}C)$. The outputs from these two paths are concatenated and passed through an upsampling layer to restore the original token resolution.

By fusing these two paths, the Weaver-Attention module provides a comprehensive feature representation with a total computational complexity of approximately $\mathcal{O}\left(\frac{N}{S}C + \left(\frac{N}{S}\right)^2 \frac{C}{4}\right)$. This is substantially more efficient than the $\mathcal{O}(N^2 C)$ of standard self-attention, enabling the Weaformer layer to achieve real-time performance.


\begin{table*}[htbp]
\centering
\caption{Performance Comparison of Different Counting Methods on FSC147}
\label{tab:counting_methods}
\resizebox{0.75\textwidth}{!}{%
\begin{tabular}{@{}clcccccc@{}}
\toprule
\textbf{Type} & \textbf{Method} & \textbf{Base model} & \textbf{Differentiable} & \multicolumn{2}{c}{\textbf{Validation Set}} & \multicolumn{2}{c}{\textbf{Test Set}} \\
\cmidrule(lr){5-6} \cmidrule(lr){7-8}
& & & & \textbf{MAE $\downarrow$} & \textbf{RMSE $\downarrow$} & \textbf{MAE $\downarrow$} & \textbf{RMSE $\downarrow$} \\
\midrule

\multirow{6}{*}{\begin{tabular}[c]{@{}l@{}} Zero-shot \\ visual prototype\end{tabular}} 
& FamNet \cite{FamNet} & ResNet-50 & -- & 32.15 & 98.75 & 32.27 & 131.46 \\
& RepRPN-C \cite{reprpnc} & ResNet-50 & -- & 29.24 & 98.11 & 26.66 & 129.11 \\
& CounTR \cite{Countr} & ResNet-50 & -- & 18.07 & 71.84 & 14.71 & 106.87 \\
& LOCA \cite{loca1} & ResNet-50 & -- & 17.43 & 54.96 & 16.22 & 103.96 \\
& RCC \cite{RCC} & ViT & -- & 17.49 & 58.81 & 17.21 & 104.53 \\
& DAVE \cite{DAVE} & ResNet-50 & $\times$ & \textbf{15.54} & \textbf{52.67} & \textbf{15.14} & \textbf{103.49} \\
\midrule

\multirow{3}{*}{\begin{tabular}[c]{@{}l@{}}Text\end{tabular}} 
& DAVE$_{\text{prm}}$ \cite{DAVE} & ResNet-50 & $\times$ & 15.48 & 52.57 & 14.90 & 103.42 \\
& CountGD \cite{CountGD} & Grounding DINO & $\times$ & 12.14 & 47.51 & 12.98 & 98.35 \\

& GroundingREC \cite{GroundingDINO} & Grounding DINO & $\times$ & 10.06 & 58.62 & \textbf{10.12} & 107.19 \\

& CAD-GD \cite{CAD-GD} & Grounding DINO & $\times$ & \textbf{9.30} & \textbf{40.96} & 10.35 & \textbf{86.88} \\
\midrule

\multirow{6}{*}{\begin{tabular}[c]{@{}l@{}}Text\end{tabular}} 
& ZSC \cite{zsc} & ResNet-50 & $\checkmark$ & 26.93 & 88.63 & 22.09 & 115.17 \\
& CLIP-Count \cite{CLIP-Count} & CLIP & $\checkmark$ & 18.79 & 61.18 & 17.78 & 106.62 \\
& CounTX \cite{CounTX} & Open CLIP & $\checkmark$ & 17.33 & 61.58 & 16.28 & 106.41 \\
& VLCounter \cite{VLCounter} & CLIP & $\checkmark$ & 18.06 & 65.13 & 17.05 & 106.16 \\ \hline
& \textbf{Ours} & YOLOE & $\checkmark$ & \textbf{12.56} & \textbf{51.25} & \textbf{13.30} & \textbf{104.63} \\
\bottomrule
\end{tabular}%
}
\end{table*}

\subsection{Loss Function Design}
RT-Counter addresses two primary learning objectives: classification and localization. therefore, we design the loss functions to optimize both tasks simultaneously.

\textbf{Notations and Matching Strategy:} Let $\mathcal{P}=\{p_i|i\in{1,...,N_{GT}}\}$ be the set of ground truth point annotations, where $p_i=(x_i,y_i)$ represents the $i$-th annotation point with coordinates $(x_i, y_i)$, and $N_{GT}$ is the total number of ground truth annotations. Through the Hungarian matching algorithm, we establish optimal one-to-one correspondences between predicted and ground truth points. Let $\hat{\mathcal{P}}_\xi=\{{\hat{p}_{\xi(j)}|j\in{1,...,N_{Grid}}}\}$ be the set of matched prediction points and $\hat{\mathcal{C}}_\xi=\{{\hat{c}_{\xi(j)}|j\in{1,...,N_{Grid}}}\}$ be the set of corresponding confidence scores for these matched predictions, where $N_{Grid}\geq N_{GT}$. Notably, the first $N_{GT}$ points in $\hat{\mathcal{P}}_\xi$ correspond one-to-one with the $N_{GT}$ ground truth annotations.

\textbf{Regression Loss:} Inspired by the repulsion loss designed in \cite{yolofacev2}, we formulate a smooth regression loss function $\mathcal{L}_{reg}$ to minimize localization errors. That is, 
\begin{align}\label{eq:reg}
	\mathcal{L}_{reg}=\frac{1}{N_{GT}}\sum_{i=1}^{N_{GT}}Smooth_{ln}\left(d(\hat{p}_{\xi(i)},p_i)\right)
\end{align}
where $d(\cdot,\cdot)$ denotes the Euclidean distance between two points, and $Smooth_{ln}(\cdot)$ is defined as
\begin{align}
	Smooth_{ln}(z)=
	\begin{cases}
		-\ln(1-z), & \text{if } z\leq \sigma \ \\
		\frac{z-\sigma}{1-\sigma}-\ln(1-\sigma), & \text{otherwise}
	\end{cases}
\end{align}
where the hyperparameter $\sigma$ is preset to control the smoothness transition.

\textbf{Classification Loss:} The classification loss $\mathcal{L}_{cls}$ employs a weighted Cross Entropy (CE) function to address the imbalance between positive and negative samples. Namely, 
\begin{equation}\label{eq:cls}
	\mathcal{L}_{cls}=-\frac{1}{N_{Grid}} \left(\sum_{i=1}^{N_{GT}}\log\hat{c}_{\xi(i)}+ \\ 0.5\sum_{i=N_{GT}+1}^{N_{Grid}}\log(1-\hat{c}_{\xi(i)})\right)
\end{equation}
where the weighting factor of 0.5 for negative samples helps mitigate the class imbalance issue commonly encountered in object counting tasks.

\textbf{Total Loss:} The overall training objective combines both losses with equal weighting. Then, we have
\begin{align}
	\mathcal{L}_{total} = \lambda_1\mathcal{L}_{reg} + \lambda_2\mathcal{L}_{cls}
\end{align}
where $\lambda_1$ and $\lambda_2$ are hyper-parameters given in advance.
\section{Experiments}
\subsection{Experimental Setup}
\textbf{Datasets:} To validate the performance of RT-Counter, we evaluate it on three widely-used counting benchmarks: FSC147 \cite{FamNet}, CARPK \cite{CarPK}, and REC-8K \cite{rec}. 
\textbf{FSC147} is a benchmark for zero-shot or few-shot counting, containing 6,135 images across 147 categories. It is designed to evaluate generalization to novel object classes with limited or zero examples, as the object categories across its training, validation, and test sets are non-overlapping.
\textbf{CARPK} is a specialized dataset for vehicle counting, comprising 1,448 aerial images of parking lots containing nearly 90,000 vehicles. It provides a domain-specific benchmark for evaluating model transferability in transportation scenarios.
\textbf{REC-8K} is a large-scale dataset for Referring Expression Counting (REC), consisting of 8,011 images and 17,122 image-RE (Referring Expression) pairs. It offers a comprehensive evaluation of counting across diverse scenarios by including a wide variety of object types and attributes such as color, material, and action.

\paragraph{Data Augmentation:} To enhance model robustness and generalization capability, we apply a comprehensive set of data augmentation techniques including random cropping, horizontal/vertical flipping, and color jittering during training.
\paragraph{Training Strategy:} We employ the AdamW optimizer with a learning rate of $1 \times 10^{-4}$ (where the backbone uses a learning rate of $1 \times 10^{-5}$) and a batch size of 8. The model is trained end-to-end without additional regularization techniques to maintain simplicity and efficiency of the training.
\paragraph{Evaluation Metrics:} We adopt standard evaluation metrics for comprehensive performance assessment, including MAE (Mean Absolute Error), RMSE (Root Mean Square Error), FPS (Frames Per Second), GFLOPs, and Params.
Both MAE and RMSE estimate the counting accuracy of models, where MAE measures the average absolute difference between predicted and ground truth counts, and RMSE provides a measure of prediction variance and penalizes larger errors more heavily. 
FPS, GFLOPs, and Params are all related to the model efficiency. 
FPS evaluates real-time inference speed for practical deployment scenarios, GFLOPs quantifies computational complexity in terms of Giga Floating-point OPerations, and Params reports the total number of model parameters, indicating model size.

\subsection{Comparison with SOTA Methods}
\paragraph{Results on FSC147:}
Table~\ref{tab:counting_methods} presents the performance comparison on the FSC147 dataset. Our RT-Counter achieves a MAE of 13.30 and a RMSE of 104.63 on the test set. 
Obviously, RT-Counter is much better compared to the best Zero-shot visual prototype-guided method DAVE, demonstrating the importance of semantic understanding in counting tasks. 
Moreover, RT-Counter achieves the best results among differentiable text-based methods, which has a 18.3\% improvement over the second-best differentiable text-based method CounTX with a MAE of 16.28.
Although GroundingREC is the best MAE of 10.12 among all compared methods, its counting performance is highly related to the high computational cost of GroundingDINO and need adaptive cropping strategy to overcome the 900 counting quota of the model \cite{CountGD}. 
Notably, the performance of our approach is close to that of non-differentiable method CountGD, while offering the advantages of gradient-based optimization.


\paragraph{Results on REC-8K:}
Table~\ref{tab:8k_perfor} shows the performance comparison on the REC-8K dataset. We can see that RT-Counter achieves a MAE of 5.72 and a RMSE of 18.19 on the test set, demonstrating competitive performance in the challenging multi-attribute counting scenario. Findings imply our significant improvements over CLIP-based methods, for example, the MAE of RT-Counter is 51.7\% lower than that of CounTX (MAE: 11.84).
Compared to GroundingDINO-based methods, our method maintains competitive results while offering the advantages of real-time inference speed and significantly reduced computational complexity.
In a nutshell, RT-Counter demonstrates strong generalization capabilities across diverse counting scenarios while maintaining its core advantages of efficiency and end-to-end differentiable training.

\begin{table}[t]
\centering
\caption{Performance Comparison on REC-8K.}
\label{tab:8k_perfor}
\resizebox{0.5\textwidth}{!}{%
\begin{tabular}{@{}lcccc@{}}
\toprule
\multirow{2}{*}{\textbf{Method}} & \multicolumn{2}{c}{\textbf{Val Set}} & \multicolumn{2}{c}{\textbf{Test Set}} \\
\cmidrule(lr){2-3} \cmidrule(lr){4-5}
& \textbf{MAE $\downarrow$} & \textbf{RMSE $\downarrow$} & \textbf{MAE $\downarrow$} & \textbf{RMSE $\downarrow$} \\
\midrule
Mean & 14.28 & 27.75 & 13.75 & 25.91 \\
ZSC$^{ResNet_{50}}$ \cite{zsc} & 14.84 & 31.30 & 14.93 & 29.72 \\
ZSC$^{SwinT}$ \cite{zsc} & 12.96 & 26.74 & 13.00 & 29.07 \\
CounTX \cite{CounTX} & 11.88 & 27.04 & 11.84 & 25.62 \\
GroundingDINO$^{SwinT}$ \cite{GroundingDINO} & 9.03 & 21.98 & 8.88 & 21.95 \\
GroundingREC$^{SwinT}$ \cite{rec} & 6.80 & 18.13 & 6.50 & 19.79 \\
CAD-GD$^{SwinT}$ \cite{CAD-GD} & 5.43 & 15.01 & 5.29 & 17.08 \\
CAD-GD$^{SwinB}$ \cite{CAD-GD} & \textbf{4.23} & \textbf{13.14} & \textbf{4.34} & \textbf{12.93} \\ \hline
ours & 5.26 & 17.26 & 5.72 & 18.19 \\
\bottomrule
\end{tabular}%
}
\end{table}

\begin{table}[htbp]
\centering
\caption{Performance Comparison on CARPK Dataset}
\label{tab:carpk_text_methods}
\begin{tabular}{@{}lcccc@{}}
\toprule
\textbf{Dataset} & \textbf{Method} & \textbf{MAE $\downarrow$} & \textbf{RMSE $\downarrow$} \\
\midrule
\multirow{4}{*}{CARPK} 
& CLIP-count \cite{CLIP-Count}   & 11.96 & 16.61 \\
& CounTX \cite{CounTX} & 6.13 & 10.87 \\
& VLCounter \cite{VLCounter}  & 6.46 & 8.68 \\
& CountGD \cite{CountGD}  & \textbf{3.83} & \textbf{5.41} \\ \hline
& Ours & 4.81 & 7.61 \\
\bottomrule
\end{tabular}
\end{table}

\paragraph{Results on CARPK:}
Table~\ref{tab:carpk_text_methods} lists the performance comparison on the CARPK dataset. CountGD achieves the best performance (MAE: 3.83, RMSE: 5.41), followed closely by RT-Counter with a MAE of 4.81 and a RMSE of 7.61. However, CountGD is unsuitable for real-time deployment due to its heavy computational cost. 
Compared to other differentiable approaches, RT-Counter significantly outperforms CLIP-Count (MAE: 11.96) and shows competitive results against both CounTX (MAE: 6.13) and VLCounter (MAE: 6.46).  This validates our method's effectiveness across different counting scenarios while maintaining the advantage of end-to-end differentiable training.


\begin{table}[htbp]
\centering
\caption{Model Efficiency Comparison on 384 $\times$ 384 image.}
\label{tab:efficiency_comparison}
\begin{tabular}{@{}lccc@{}}
\toprule
\textbf{Method} & \textbf{Parameters $\downarrow$} & \textbf{FPS $\uparrow$} & \textbf{GFLOPs $\downarrow$} \\
\midrule
CountGD \cite{CountGD} & 213M & 4.6 & 98.96 \\
CLIP-Count \cite{CLIP-Count} & 168M & 12.15 & 112.70 \\
CounTX \cite{CounTX} & 161M & 15.11 & 43.88 \\
DAVE$_{prm}$ \cite{DAVE} & 150M & 3.01 & 170.77 \\
\textbf{Ours} & \textbf{38M} & \textbf{112.48} & \textbf{21.37} \\
\bottomrule
\end{tabular}
\end{table}

\paragraph{Model Efficiency Comparison:}
Table~\ref{tab:efficiency_comparison} demonstrates the superior efficiency of our RT-Counter framework across multiple computational metrics. Our method achieves remarkable efficiency improvements while maintaining competitive counting accuracy.

RT-Counter achieves significant model compression with only 38M parameters, representing a 74.6\% reduction compared to the smallest existing method DAVE$_{prm}$ (150M). This substantial parameter reduction directly translates to reduced memory requirements and faster model loading times. The inference speed improvements are particularly striking, with RT-Counter achieving 112.48 FPS compared to the fastest existing method CounTX at 15.11 FPS, representing a 7.4$\times$ speedup that makes it highly suitable for real-time applications.

Our method also demonstrates exceptional computational efficiency with only 21.37 GFLOPs, achieving 2.1$\times$ better efficiency than CounTX (43.88) and 4.6$\times$ better efficiency than CountGD (98.96). These comprehensive efficiency gains across parameters, inference speed, and computational complexity validate that our lightweight architecture design successfully balances high counting accuracy with practical real-time deployment requirements.


\begin{table}[htbp]
\centering
\caption{Ablation Study Results on FSC147.}
\label{tab:ablation_study}
\begin{tabular}{lcccc}
\toprule
\multicolumn{1}{c}{\multirow{2}{*}{\textbf{Method}}} & \multicolumn{2}{c}{\textbf{FSC-147 Val}} & \multicolumn{2}{c}{\textbf{FSC-147 Test}} \\
\cmidrule(lr){2-3} \cmidrule(lr){4-5}
 & MAE & RMSE & MAE & RMSE \\
\midrule
Baseline (zero-shot) & 43.12 & 139.06 & 33.28 & 142.87 \\
Baseline (Point) & 30.25 & 87.63 & 30.55 & 109.08 \\
\midrule
Ours w/o FE & 17.52 & 68.19 & 22.98 & 108.32 \\
Ours w/o VPT & 14.66 & 56.41 & 16.09 & 110.84 \\
\midrule
\textbf{Ours (Full Model)} & \textbf{12.56} & \textbf{51.25} & \textbf{13.30} & \textbf{104.63} \\
\bottomrule
\end{tabular}
\end{table}

\begin{figure}
    \centering
    \includegraphics[width=0.9\linewidth]{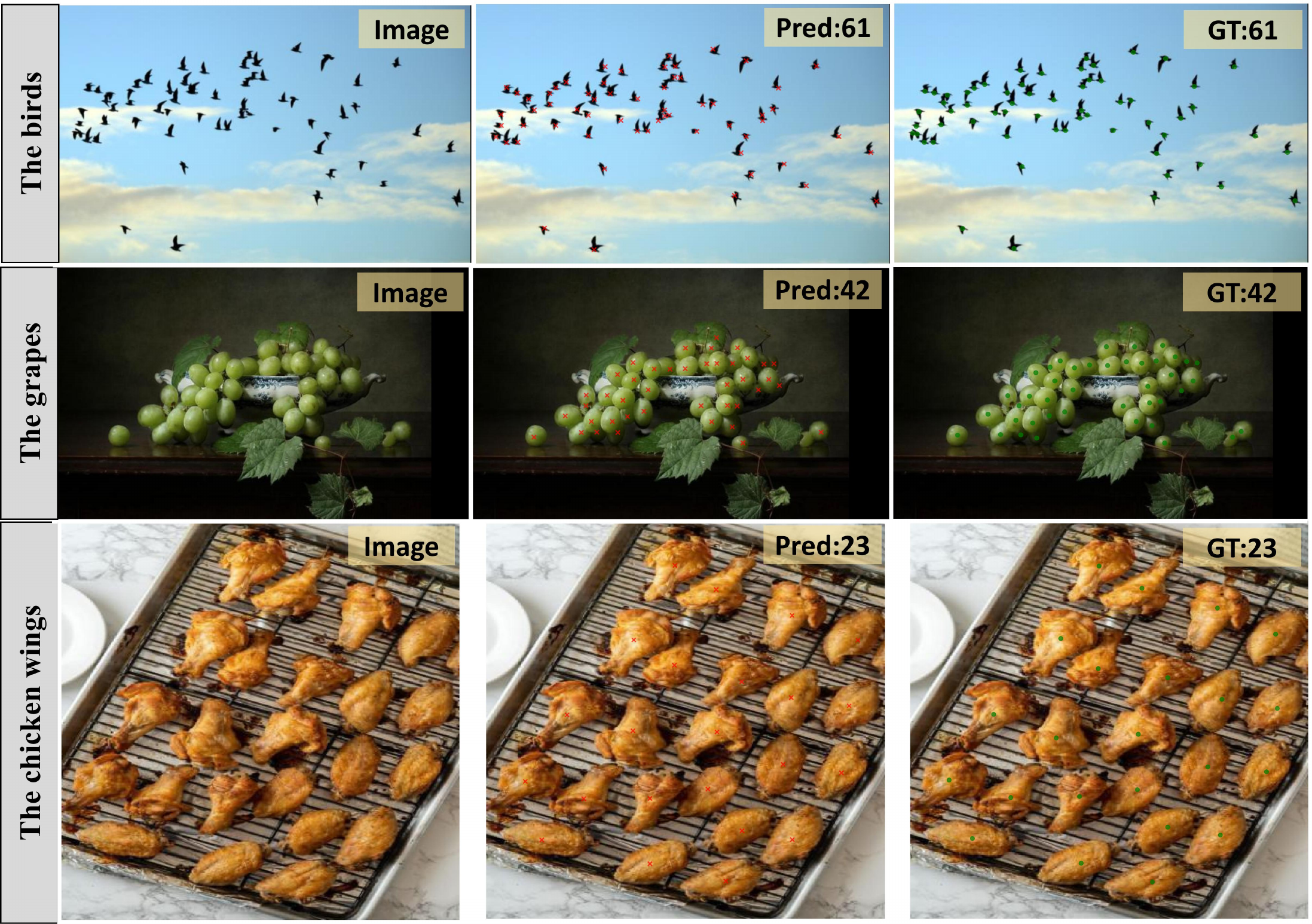}
    \caption{The counting results on FSC147.}
    \label{fig:result}
\end{figure}

\paragraph{Qualitative Analysis:}
Figure~\ref{fig:result} provides qualitative results demonstrating the good counting performance of RT-Counter across diverse scenarios.
In the top row of Figure~\ref{fig:result}, RT-Counter accurately counts birds in flight (Pred: 61, GT: 61), which shows its excellent performance on small, densely distributed objects against complex backgrounds. The middle row gives the precise grape counting (Pred: 42, GT: 42) of RT-Counter, illustrating its capability to handle overlapping objects with similar textures and colors. The bottom row presents the chicken wing counting on a cooking rack (Pred: 23, GT: 23), showcasing the model's accurate detection of food items with varying orientations and partial occlusions.

These results highlight RT-Counter's robust generalization across different object categories, scales, and environmental conditions. The model successfully handles challenging scenarios including object overlapping, background clutters, and various lighting conditions. 

\subsection{Ablation Studies}
Table~\ref{tab:ablation_study} presents a comprehensive ablation study demonstrating the contribution of main components in our RT-Counter framework. While point supervision provides a strong baseline over a zero-shot approach, removing either the Feature Enhancer (Ours w/o FE) or the Visual Prototype Textualization module (Ours w/o VPT) leads to significant performance degradation. Notably, removing the VPT module causes the most substantial drop in accuracy (test MAE from 13.30 to 16.09), highlighting it as the most critical contributor to our model's success. The superior performance of our full model confirms that all components work synergistically to achieve the final result.

\begin{table}[htbp]
\centering
\caption{Comparison of VPT Integration on FSC147 Performance.}
\label{tab:vpt_comparison}
\begin{tabular}{lcccc}
\toprule
\multicolumn{1}{c}{\multirow{2}{*}{\textbf{Method}}} & \multicolumn{2}{c}{\textbf{FSC-147 Val}} & \multicolumn{2}{c}{\textbf{FSC-147 Test}} \\
\cmidrule(lr){2-3} \cmidrule(lr){4-5}
 & MAE & RMSE & MAE & RMSE \\
\midrule
CounTX & 17.33 & 61.58 & 16.28 & 106.41 \\
CounTX + VPT & \textbf{15.85} & \textbf{59.73} & \textbf{15.39} & \textbf{105.81} \\
\midrule
Ours w/o VPT & 14.66 & 56.41 & 16.09 & 110.84 \\
\textbf{Ours (Full Model)} & \textbf{12.56} & \textbf{51.25} & \textbf{13.30} & \textbf{104.63} \\
\bottomrule
\end{tabular}
\end{table}

Beyond its effectiveness within our own framework (as shown in Table~\ref{tab:ablation_study}), we further validate the generalizability and plug-and-play nature of our VPT module through a cross-method integration experiment. The results are presented in Table~\ref{tab:vpt_comparison}.

We integrated the VPT module into a recent baseline, CounTX \cite{CounTX}. The results show a clear performance boost, with the test MAE improving from 16.28 to 15.39, a notable reduction of 0.89. This experiment confirms that VPT is not merely tailored to our specific architecture but is a portable component that can effectively enhance other counting frameworks. For a direct comparison, the table also shows that VPT brings an even more substantial improvement to our own model (a 2.79 MAE reduction on the test set). Collectively, these results demonstrate that our VPT module is a robust and effective component.

\begin{table}[t]
\centering
\caption{Ablation study of the key components in our Weaformer.}
\label{tab:ablation_Weaformer}
\begin{tabular}{lcccc}
\toprule
\multicolumn{1}{c}{\multirow{2}{*}{\textbf{Method}}} & \multicolumn{2}{c}{\textbf{FSC-147 Val}} & \multicolumn{2}{c}{\textbf{FSC-147 Test}} \\
\cmidrule(lr){2-3} \cmidrule(lr){4-5}
 & MAE & RMSE & MAE & RMSE \\
\midrule
w/o Downsampling & 13.64 & 55.57 & \textbf{13.17} & 105.98 \\
w/o ConvAttention & 13.10 & 52.49 & 13.94 & 111.52 \\
w/o Self-Attention & 13.59 & \textbf{50.93} & 14.42 & 115.96 \\
\midrule
\textbf{Ours (Full Model)} & \textbf{12.56} & 51.25 & 13.30 & \textbf{104.63} \\
\bottomrule
\end{tabular}
\end{table}
To validate our design, we conduct an ablation study on the key components of the Weaformer layer, with results shown in Table~\ref{tab:ablation_Weaformer}. Removing the initial downsampling step (w/o Downsampling) leads to a clear performance drop, highlighting its importance for efficiency and feature refinement. Disabling either the local path (w/o ConvAttention) or the global path (w/o Self-Attention) results in a significant degradation in accuracy. Notably, removing the global self-attention causes the most severe performance drop, with the test MAE increasing from 13.30 to 14.42, which underscores the critical role of modeling long-range dependencies. Our full model consistently outperforms all ablated variants, confirming that the local and global pathways are complementary and essential for achieving optimal performance.

\section{Conclusion}
In this work, we present RT-Counter, a real-time text-guided counting framework that successfully reconciles high accuracy with computational efficiency. This is achieved through two primary contributions: (1) the novel VPT module, which effectively bridges the modality gap by enriching visual prototypes with textual semantics; (2) the highly efficient Weaformer layer, which processes multi-scale information by weaving together local and global features at a fraction of the standard computational cost.
Our findings demonstrate that high-performance open-vocabulary counting is practical for real-world, time-sensitive applications, setting a new standard for future research.


{
    \small
    \bibliographystyle{ieeenat_fullname}
    \bibliography{main}
}


\end{document}